# Curriculum-based Sensing Reduction in Simulation to Real-World Transfer for In-hand Manipulation

Lingfeng Tao[#], Jiucai Zhang[^], Qiaojie Zheng[*], and Xiaoli Zhang[*], *Senior Member, IEEE*

*Abstract* — **Simulation to Real-World Transfer allows affordable and fast training of learning-based robots for manipulation tasks using Deep Reinforcement Learning methods. Currently, Sim2Real uses Asymmetric Actor-Critic approaches to reduce the rich idealized features in simulation to the accessible ones in the real world. However, the feature reduction from the simulation to the real world is conducted through an empirically defined one-step curtail. Small feature reduction does not sufficiently remove the actor's features, which may still cause difficulty setting up the physical system, while large feature reduction may cause difficulty and inefficiency in training. To address this issue, we proposed Curriculum-based Sensing Reduction to enable the actor to start with the same rich feature space as the critic and then get rid of the hard-to-extract features step-by-step for higher training performance and better adaptation for real-world feature space. The reduced features are replaced with random signals from a Deep Random Generator to remove the dependency between the output and the removed features and avoid creating new dependencies. The methods are evaluated on the Allegro robot hand in a real-world in-hand manipulation task. The results show that our methods have faster training and higher task performance than baselines and can solve real-world tasks when selected tactile features are reduced.**

## I. INTRODUCTION

Dexterous in-hand manipulation is one of the essential functions for robots in human-robot interaction [1], intelligent manufacturing [2], telemanipulation [3], and assisted living [4], but it is also hard to solve due to the high degrees of freedom (DoFs) in control space and the complex interaction with the object. Deep Reinforcement Learning (DRL) [5] has shown its abilities in recent research [6-8] to solve dexterous in-hand manipulation tasks thanks to its learning capability, which enables the robot to find a control policy by interacting with the environment through exploration and exploitation.

Recent literature uses Simulation to Real-world (Sim2Real) [9] transfer, which trains the DRL policy in the simulated environment and then transfers the policy to the real robot to complete the same task. The benefit of using Sim2Real is that the simulation platform can be easily customized to recreate the real-world environment, reducing the implementation effort. The training can be accelerated with multi-thread and parallel training setups [6]. Most importantly, the simulation environment can provide more explicit information [10] that is hard to extract in the real world, such as tactile, depth, and thermal sensing, to expand the feature space of the DRL policy and increase task performance.

This material is based on work supported by the US NSF under grant 1652454 and 2114464.

[#]L. Tao is with Oklahoma State University, 563 Engineering North, Stillwater, OK, 74075 (email: lingfeng.tao@okstate.edu).

[^]J. Zhang is with the GAC R&D Center Silicon Valley, Sunnyvale, CA 94085 USA (e-mail: zhangjiucai@gmail.com).

[*]Q. Zheng, and X. Zhang are with Colorado School of Mines, Intelligent Robotics and Systems Lab, 1500 Illinois St, Golden, CO 80401 (e-mail: tao@mines.edu, xlzhang@mines.edu).

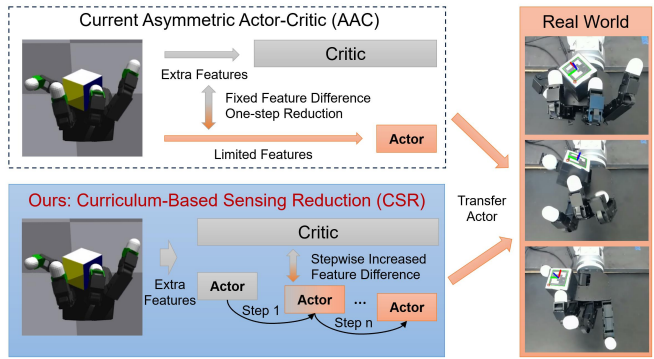

Fig. 1. Compare current AAC with our CSR method. The AAC approach is a fixed, one-step feature reduction that cannot maximize the benefit from the rich feature space in the simulation environment. The proposed CSR method gradually reduces the feature space for the actor from the same as the critic to a smaller feature space that is suitable for the real world for better performance and easier implementation in the real world.

Although the rich information in simulation can significantly improve policy performance, a side-effect is that setting up the same information space in the real world is challenging and expensive. The more information used in simulation, the more effort and cost to extract the information in the real world will be needed. For example, a single BioTac [11] tactile sensor attached to the fingertip for the Allegro hand [12] costs $15k. In the OpenAI Gym simulation platform [13], the Shadow hand can have 92 tactile sensors that cover the whole hand [14], which is impractical in the real world. To reduce implementation effort in the real world, high-dimensional inputs like images are used [6], but the DRL model needs extra convolutional layers to extract information from the sparse data [15], increasing the data required and the training burden. Further, high-dimensional data are usually highly dependent on environmental conditions. When transferred to the real world, any change to the light condition, color, and camera setup will significantly affect the policy performance [6].

To utilize extra information in the simulation, researchers developed the Asymmetric Actor-Critic (AAC) [16] method based on the conventional actor-critic method. AAC trains a critic to approximate the value function to predict the state information's potential reward and the actor policy's control output. When training AAC for Sim2Real in simulation, the critic can observe explicit information (e.g., joint, position, tactile) and help the training of the actor, who can only observe ambiguous information (e.g., image) to accommodate the real world. After the training, only the actor is transferred, requiring less setup effort in the real world. Although AAC allows the feature reduction from the rich idealized features in simulation to the accessible ones in the real world, it still has significant shortcomings that lead to a non-adaptable actor-critic feature gap and suboptimal actor performance. This shortcoming stems from the empirically defined one-step curtail that reduces features from the simulation to the real world [6-7]. Small feature reduction does not sufficiently remove the actor's features, which still causes difficulty setting up the physical system, while large feature reduction may cause difficulty and inefficiency in policy training. Thus, in the current AAC, the actor needs to compromise to balance the feature extraction effort and the learning performance, resulting in the incapability of maximizing the benefit from the rich information in the simulation.

In this work, our rationale is that the actor should start with the same rich feature space as the critic, then get rid of the hard-to-extract features **step-by-step (Fig. 1).** Such a strategy enables higher training performance with the rich information at the start of training and gradual adaptation for real-world feature space. To achieve this goal, there are two questions to answer. First, how to decide which features need to be reduced in each step? Second, how to reduce features during the training without sacrificing training stability? To address the first question, we proposed the Curriculum-based Sensing Reduction (CSR) method. To achieve state space reduction, CSR evaluates the importance of the target features during the early training period based on expert-defined feature importance measures and builds a sensing reduction curriculum that specifies which features are removed at each step, with an objective that avoids influence on the task performance. To address the second question, we developed the Deep Random Generator (DRG) using a deep random neural network [19] to generate random signals to replace the reduced feature signal. As the actor adapts to the random signal, DRG will remove the dependency between the output and the reduced features and avoid creating new dependencies. In summary, the contributions of this work are:

1) Developed CSR to generate the stepwise feature reduction curriculum based on the feature importance to help the control policy gradually adapt to the limited feature space.
2) Developed the DRG method to generate random signals that replace the signal from the reduced features, remove the dependency between the output and the reduced features, and avoid creating new dependencies.
3) Incorporated low dimensional input to reduce the model size and implementation effort in the real world by improving data explicitly and learning efficiency.
4) Validated CSR and DRG on the Nvidia Isaac Gym robotics simulation platform in an in-hand manipulation task using an Allegro hand. The trained policy is transferred to the physical Allegro hand and tested in the task performance in real-world experiments in multiple control scenarios.

## II. RELATED WORK

### A. Learning-based In-hand Manipulation

The rapid development of dexterous robotic hands has provided hardware foundations, such as the Allegro hand [12], an anthropomorphic robotic hand with 16 DoFs in which all joints are controllable. Tactile sensors like temperature [20], Hall effect [21], and electroactive polymeric [22] are developed to improve the fidelity of the robot's observation space. With the readiness of robot hardware, researchers have been putting efforts into developing generalizable and adaptable control methods for in-hand manipulation applications. DRL methods have demonstrated their capability to handle in-hand manipulation tasks [6-8]. The OpenAI Gym [13] toolkit implements challenging in-hand manipulation tasks [23]. Nvidia released their Isaac Gym platform [24] that can train the DRL policy on GPU with much faster simulation and a more realistic environment thanks to the CUDA and Physx engine.

### B. Sim2Real for In-hand Manipulation

Recent literature focuses on deploying the DRL agent trained in simulation to the physical robot hand to complete real-world tasks. Domain randomization [25] is applied for in-hand manipulation to sensing, actuation, and appearance to improve the policy adaptability to noise and disturbance in the real world. To solve the observation gap, the key technology of Sim2Real for in-hand manipulation is AAC. The first AAC for Sim2Real of the robot arm was proposed in [16], where the critic observes explicit information to help the actor who can only observe the image. AAC was then adopted by much Sim2Real research, such as controlling a robot arm to open drawers [25], manipulating deformable objects with a gripper [26], and learning to crawl with a soft robot [27]. AAC has been adopted for in-hand manipulation to control a Shadow hand to solve a Rubik's cube or rotate a block to a target pose [7]. However, in the current AAC, the feature space of the actor and critic is still pre-determined and fixed, researchers must make compromises to balance between learning efficiency and implementation effort in the real world.

### C. Curriculum Learning for Improving Learning Efficiency

Recent research proposes the Curriculum Learning (CL) method to improve the learning efficiency and performance of RL training. CL trains the DRL policy on a series of easier tasks toward the target tasks with a selected sequence [18]. In [28], a reverse curriculum generation method was proposed to gradually learn to reach the goal from a set of start states increasingly far from the goal, which leads to efficient training on goal-oriented tasks. A graph-based curriculum representation was proposed in [29] to automatically decide the fixed learning sequence of the objectives within the time threshold. A curriculum was implemented in [30] to continuously update the reward function during training. Current CL approaches are used in task design and reward design domains. These approaches inspired the authors to adopt the gradual learning strategy and introduce it to the feature space domain for the first time in in-hand manipulation tasks.

## III. METHODOLOGY

This section introduces the modeling of in-hand manipulation in III. A. The development of CSR is explained in III. B. The development of the DRG is explained in III. C.

### A. Modeling and Representation

In this work, the in-hand manipulation task is modeled as a Markov Decision Process [31], which is described as a tuple $[A, S, T, R, \gamma]$, where $A$ is a set of actions, $S$ is a set of states. and $T(s_{t+1}|s_t, a_t)$ is the state transition probability to describe the probability of action $a$ in state $s$ at time $t$ leading to state $s_{t+1}$ as time $t + 1$. In this work, the environment is assumed deterministic, so $T = 1$. $r = R(s_{t+1}|s_t, a_t)$ is the reward received after the transition from state $s_t$ to state $s_{t+1}$. $\gamma$ is a discount factor. A policy $\pi(s)$ specifies the action for state $s$. A Proximal Policy Optimization algorithm [32] is

adopted to solve the in-hand manipulation task by approximating the DRL policy with the objective:
$$L = \hat{E}_{a_t, x_t \in D}[log\pi_\theta(a_t|x_t)\hat{A}_t] \quad (1)$$
The policy $\pi_\theta$ is represented with a Deep Neural Network as $\pi(s, \theta)$, where $\theta$ is the network parameters. $\hat{A}_t$ is an estimator of the advantage function at timestep $t$. $D = \{a_t, x_t, r\}$ is the set of past transitions. The observable state is denoted as $x \in S$, including the positions, velocities, and torques of the robot's joints, the Cartesian position, and rotation of the object represented by a quaternion as its linear and angular velocities, the target position, and the tactile information. As a validation test case, tactile information is the target to be removed in this work. The action and state spaces are normalized to -1 to 1 for stable training.

*B. Curriculum-based Sensing Reduction*

Specifically, for in-hand manipulation tasks, explicit observation enables both the actor and critic to increase the task performance with low dimensional feature space, including the information from the robot joints reading, object position, and tactile sensors. Then, the easy-to-get features remain, such as the joint information, which can be recorded from encoders, and the object position, which can be tracked with a camera. The hard-to-get features like tactile information will be reduced stepwise during the training so the actor can gradually adapt to the limited information and maintain the performance. To generate a curriculum that helps the actor adapt gradually, the first step is to identify which features need to be reduced first and which need to be reduced later. The reduction sequence is calculated by evaluating the feature's importance during the training. We define a criterion based on feature's impact on task performance. A feature is deemed important when it greatly impacts task performance when reduced, and vice versa. However, it is hard to directly measure such impact by removing the features one by one during the training. To address this issue, evaluation metrics that relate to task performance based on expert interpretations can be empirically defined. For example, possible metrics are that the tactile sensors on the fingertips are more important than those on the palm or the tactile sensors on the edge of the hand are more important than the sensors on the inner side.

The online metrics that evaluate features during manipulation can be defined, which may give a more accurate comparison. This work uses an online metric based on the activation number for each tactile sensor. The sensor with a higher activation number is more important, which means the policy relied more on it when interacting with the object during the in-hand manipulation process. With the defined metric, the features can be ranked based on the importance level, and the curriculum can be generated to reduce the features starting from the least important one. The features set in the actor's observation are denoted as $x \leftarrow \{f_1, \dots, f_n\}$, where $n$ is the number of features, and each $f$ is the sensor reading. The metric to evaluate the feature is calculated is denoted as $g(f)$. For a $m$ step curriculum $C = \{c_1, \dots, c_m\}$, the features that will be reduced in each step are extracted by
$$c_m \leftarrow argmin_i[g(f_1), \dots, g(f_n)] \quad \forall i \in c_m \quad (2)$$
where $i$ is the number of features removed for each step. In this work, $i$ and $m$ are empirically defined. During training, each step in the curriculum is triggered based on a performance threshold $\tau$, determined based on the cumulative episode reward. If the episode reward is higher than $\tau$, CSR will be activated to reduce features according to the importance measure.

*C. Deep Random Generator for Decremental Feature Curriculum Learning*

Once the CSR selects the target features, the next step is to reduce these features during the training. The challenge lies in reducing the number of features without altering the model structures or compromising the learning mechanism. Because the training of the DRL model approximates the non-linear relation between the input and output, where the output becomes bonded and dependent on the input as the model updates, thus, reducing the part of the input features will break the already learned dependency and affect the task performance. Literature [17] tried directly replacing the feature signal with a constant scalar value like zeros. However, we argue that this approach will not completely remove the dependency and may create a new dependency between the model output and the constant value, resulting in performance reduction. To address this issue, we develop DRG (Fig. 2) to generate random signals inside the DRL model to replace the reduced feature signal. DRG is inspired by network randomization [19] methods, originally developed to improve the generalizability of the DRL policies by randomizing the observation signal with a deep random layer while keeping all the feature information with no bias. DRG adopts the deep random layer approach to generate random signals that completely replace the reduced feature signals. DRG consists of two steps of signal randomization. First, the reduced features $f$ are randomly sampled from a normal distribution:
$$f \sim \mathcal{N}(\delta, \sigma^2) \quad (3)$$
where $\mathcal{N}$ is the normal distribution, $\delta$ is the mean, and $\sigma$ is the deviation. Combined with the remaining signals, the new randomized observation $\hat{x}_t = \mathbb{F}(x_t, \varphi)$ will be processed with a deep random layer, whose weights $\varphi$ is initialized at the start of every training epoch with a Xavier normal distribution [36]:
$$P(\varphi) = \alpha(\varphi = I) + (1-\alpha)\mathcal{N}\left(0; \sqrt{\frac{2}{n_{in} + n_{out}}}\right) \quad (4)$$
where $I$ is an identity kernel, $\alpha \in [0, 1]$ is a weighting parameter, and $n_{in}, n_{out}$ are the number of input and output channels. The Xavier normal distribution is used for randomization inside the DRL model because it maintains the variance of the input observation $x$ and the randomized input $\hat{x}$. Then the DRL objective function becomes:
$$L' = \hat{E}_{a_t, x_t \in D}[log\pi_\theta(a_t|\mathbb{F}(x_t, \varphi))\hat{A}_t] \quad (5)$$

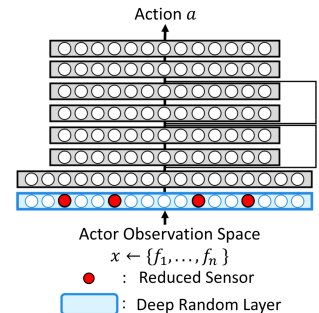

Fig. 2. An illustration of DRG. The blue layer is the deep random layer, which is initialized with a Xavier normal distribution. The red nodes are the target features, whose signal will be replaced by randomized signals to remove the existing input-output dependency and avoid creating new dependency.

With the two-stepped randomization, we expected DRG to remove the dependency between the output and the reduced features and avoid creating new dependencies. Combined with CSR, the training process is shown in Algorithm 1.

## IV. EXPERIMENTS

### A. Task Design

The CSR and DRG will be evaluated in a simulated environment for training, and the policy will be transferred and tested on the physical robot to conduct real-world in-hand manipulation tasks. The Allegro hand environment uses 13 tactile sensors from the Nvidia Isaac Gym platform [24]. The Allegro hand is a 4-finger robot hand with 16 DoFs. The in-hand manipulation task (Fig. 3) is designed to rotate a block placed on the Allegro hand's palm. The task is manipulating the block around the Z-axis to achieve the target position that is randomly generated. Four tactile sensors are attached to every finger, and one tactile sensor is on the palm. The same domain randomization in [24] is applied to the simulation. The observation space includes the joint position, velocity, torque, object position, quaternion, target quaternion, and tactile information. In the simulation, the observation is generated from the simulation API. In the real world, the joint information is extracted from the ROS node, and the object position and rotation are captured with an Aruco code [33] attached to the top of the object. A Logitech C930e webcam is used to track the Aruco code and estimate the object position and orientation at a speed of 30Hz. Kalman filter [34] is used to denoise the signal due to the detection loss during the manipulation process. The training target is achieving the desired task performance while removing the tactile sensors (Table I). This means the actor policy will start with 75 features in observation, then reduce to 62 features once the training is finished.

During training, a goal is considered achieved if the difference in the rotation is less than 0.1 rad. During testing, the criterion changes to 0.4 rad. Such a setting helps the policy achieve higher performance. In the simulation, the environment is simulated at 60Hz to capture subtle dynamics. The actor controls the Allegro hand at 30Hz to accommodate the speed limit on the physical hand. The PC hardware for training includes an Intel 12900K, a Nvidia RTX3080ti, and 64 GB of RAM. Most hyperparameters are from [24], but with changes to the number of environments to 8192, total epoch to 20000, and mini-batch size to 65536.

### B. Evaluation Metrics

Based on the benchmark results from Nvidia Isaac Gym, the curriculum trigger threshold is set at $\tau = 2500$. The following curriculums were designed for comparison. The number of features that will be reduced at each step is empirically defined, and the specific features will be selected with the importance measure:

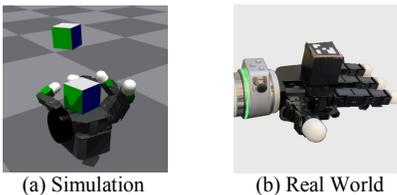

(a) Simulation  (b) Real World

Fig. 3. (a) The simulation setup and (b) Real-world setup. The tasks is to rotate the block around Z-axix in-hand to reach the target position that is randomly generated. A single top-down camera is used in real-world to detect the postion and rotation angle of the object.

**Algorithm 1 CSR**

**Initialize** critic $Q^\mu$, actor $\mu_\theta$, replay buffer $D$, importance measure $g$, performance threshold $\tau$, max-episode-length $T$, max-episode $k$.
1: **for** episode = 1 to $k$ **do**
2:    Initialize state $x$, initialize curriculum index $m = 1$
3:    **for** $t = 1$ to $T$ **do**
4:       Selection action $a_t = \mu_\theta(x_t) + \mathcal{N}$
5:       Save transition $(x, a, r, x')$ to $D$
6:       Calculate $g(f)$ for all target sensors
7:    **end for**
8:    Sample transition $(x, a, r, x')$ from $D$, calculate $\sum r$
9:    **if** $\sum r > \tau$ **do**
10:      Generate curriculum $c_m$ with eq. (1)
11:      Replace selected sensor signal with DRG
12:      $m = m + 1$
13:    **end if**
14:    Update actor $\mu_\theta$
15:    Update target network parameters $\theta'$
16: **end for**

1) CSR (3 steps) + DRG: This curriculum contains three steps. The first step reduces 4 tactile features. The second step reduces 4 tactile features. The last step reduces 3 tactile features. The signal will be processed with DRG.
2) CSR (2 steps) + DRG: It contains two steps. The first step reduces 7 tactile features. The second step reduces 6 tactile features. The signal will be processed with DRG.
3) CSR (2 steps): The curriculum is the same as (2), but the signal will be replaced with zeros instead of DRG.
4) Asymmetric Actor-Critic (one-step curtail): AAC is considered a one-step curtail baseline where the actor starts the training without any tactile features.

It should be noted that the final actor feature space for all configurations is the same for a fair comparison. The comparison focuses on one-step curtail vs. multi-step curriculums based on the same overall feature reduction. For multi-step curriculums, our method determines when and which to remove features. Comparing 1) and 2) evaluate how the number of steps will affect the training process. Comparing 2) to 3) evaluate the effectiveness of DRG. Comparing 1) to 4) evaluate CSR's improvements to the traditional AAC. The following evaluation metrics are used:

**Learning efficiency**: The learning efficiency in a specified period assesses how the curriculum improves the training.

**Task performance:** The task performance is compared in both simulations and real-world for a comprehensive comparison. We consider the success rate as the evaluation metric. The success rate is the percentage of successful cases in a testing set with 100 trials. Each trial has randomly generated initial and target poses. The testing set is reused in all evaluations for a reproducible comparison. The episode length is set to 10 seconds for easier implementation in the real world. During the manipulation process, a new target position will be chosen once the last target is achieved. The robot can continuously rotate the object with less manual reset in the real world.

TABLE I. OBSERVATION SPACE AND TARGET TACTILE SENSORS

| | Type | Length | | |
|---|---|---|---|---|
| Joint | Position | 16 | Lower phalanx of the fingers | 1 Sensor x 3 |
| | Velocity | 16 | | |
| | Torque | 16 | Middle phalanxes of the fingers | 1 Sensor x 3 |
| Object | Quaternion | 4 | | |
| | Angular velocity | 3 | Tip phalanxes of the fingers | 1 Sensor x 3 |
| Target | Position | 3 | Thumb phalanxes | 1 Sensor x 3 |
| | Quaternion | 4 | | |
| Tactile sensors | | 13 | Palm | 1 Sensor x 1 |
| Total observation | | 75 | All tactile sensors | 13 Sensors |

**Manipulation behavior**: How the trained policy manipulates the object is another critical factor that affects task performance in the real world. The preliminary test shows that the operation frequency is closely related to the manipulation behavior. Also, the physical hand usually cannot achieve the same operating frequency as the simulation due to the motor speed, collision, and stochastic dynamics. We will vary the control frequency in the real world to study how different operation speeds affect manipulation behaviors and task performance. In practice, we test the control frequency at the original speed (30Hz) and slowdown three times (10Hz) and six times (5Hz). We count the number of success cases during the 30s of consecutive rotations to measure the task performance under different control frequencies. The performance will be compared with the literature [6][35]. The manipulation behavior will be analyzed in the next section.

## V. RESULTS AND DISCUSSION

### A. Training Process

The training process is shown in Fig. 4. Overall, the 2-step CSR + DRG (blue) achieved the best performance than others, proving the effectiveness of the proposed method. Comparing the 2-step CSR + DRG (blue) and 3-step CSR + DRG (red), both methods had similar learning efficiency at the start of the training and triggered the feature reduction at a similar time. After the first CSR step, 2-step CSR + DRG had a worse performance reduction than 3-step CSR because it reduced more features. Interestingly, 2-step CSR + DRG caught up with the 3-step CSR + DRG and triggered the second curriculum step at a similar time. The potential explanation is that the feature importance measure correctly dropped the unnecessary features, so 2-step CSR + DRG relied less on those reduced features. After the second CSR step, the 2-step CSR + DRG completed the feature reduction and removed all tactile features. However, contrary to the first step, the 2-step CSR + DRG had less performance reduction than the 3-step CSR + DRG. A possible reason is that in late training, the 3-step CSR + DRG depended more on the tactile features, causing a higher impact on the performance at the second curriculum step. For the same reason, at the third curriculum step, the 3-step CSR + DRG had another performance reduction and didn't show an increase at the end of training. It may catch up with further training, but we didn't show it since we want to compare the learning efficiency under the same number of training epochs.

The 2-step CSR without DRG (yellow) spent more time catching up after the first curriculum step and struggled with low performance after the second curriculum step. This means the zeros in the replaced signal affect the performance and emphasize the importance of DRG. The baseline AAC also faced a performance drop in early training. A possible reason is over-fitting or local suboptimality. Then its performance slowly increased in later training but still had a lower performance at the end. A necessary interpretation is that the number of curriculum steps is one important factor affecting learning speed. Another important message is that the target features should be reduced at an appropriate time to benefit the policy training and avoid building strong dependencies.

The reduced tactile features at each curriculum step are shown in Fig. 4. The remaining features are in orange, and the removed features are marked in blue. A shared phenomenon across all curricula is that the features close to the lower phalanx of the fingers are reduced first. The reason is that the policy learned a manipulation behavior that relies more on fingertips. The lower phalanx tactile features are triggered less than the middle and fingertip features and, therefore, less important. As a result, those features are reduced first.

### B. Performance Evaluation

Table II shows the results of task performance in both simulation and real world. Corresponding to the training results, CSR (2 steps) + DRG achieved the highest success rate in all tests. The CSR (3 steps) + DRG came to the second due to the impact on task performance after the third curriculum step. Without DRG, CSR (2 steps) converged to a sub-optimal policy, which performs poorly in all tests. The baseline AAC performs similarly to the CSR (2 steps), but it may catch up as its performance is still increasing at the end of the training.

The control frequency significantly impacts the task performance. For example, although the CSR (2 steps) + DRG achieved a 96% success rate in simulation at 30Hz, the performance dropped 64% to 32% when keeping the same control frequency in the real world. If we slow down the speed 3 times to 10Hz, the success rate gets higher than 30Hz. However, the performance worsens when we further slowdown to 5Hz. This behavior is consistent across all configurations. To explain why, we plot the actual (red) and the goal (blue) position for the first joint of the thumb under different frequencies in Fig. 5. Due to the different time scales, we focus on the following ability: whether the actual joint matches the goal position. The results show that at 10Hz, the robot has the best following ability. At 30Hz, the command is too fast to follow because the speed of the servo motor is limited. At 5Hz, the robot cannot reach the target in time when the command is too far from the current position. This is because the control designed in the Allegro hand always tries to achieve the goal in a specified time for smooth motion, but it cannot reach the goal when it is too far. We expect the performance to improve by developing controllers with fine-tuned acceleration curves, which will be studied in the future.

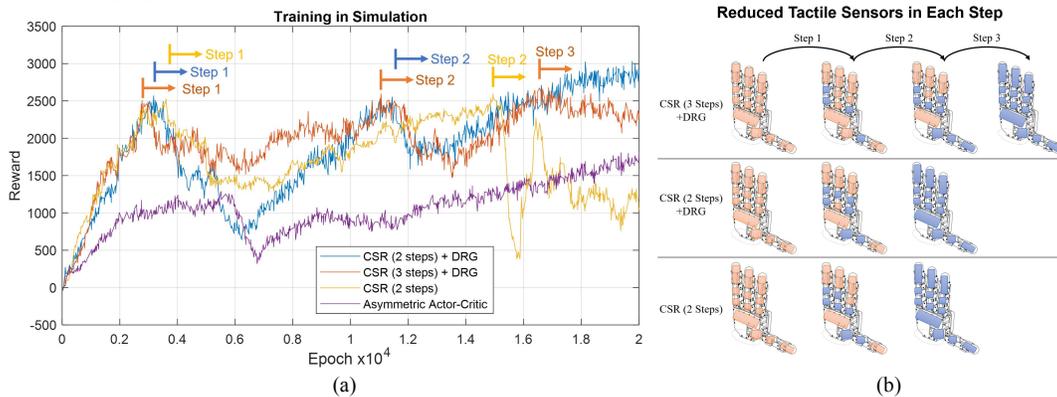

Fig. 4. (a) The 2-step CSR + DRG (blue) achieved the highest task performance in the designed training period. The 2-step CSR without DRG (yellow) has difficulty in increasing performance after the first curriculum step and dropped performance after the second step. The 3-step CSR + DRG (red) has a negative impact on the performance at the last curriculum step. The baseline AAC (purple) has a much lower task performance and less learning efficiency at the end of training. (b) The reduced tactile features in each curriculum step (AAC is not shown since its actor starts and ends with no tactile features). The reduced features are marked in blue. It shows that the sensors close to the lower phalanx of the fingers are reduced first because the policy mainly relies on the fingertips.

TABLE II. TASK SUCCESS RATE OVER 100 TRIALS

| Config | Sim-30Hz | Real-30Hz | Real-10Hz | Real-5Hz |
|---|---|---|---|---|
| CSR (2 steps)+DRG | **96%** | **32%** | **58%** | **37%** |
| CSR (3 steps)+DRG | 92% | 27% | 49% | 33% |
| CSR (2 steps) | 75% | 18% | 23% | 21% |
| AAC | 84% | 22% | 34% | 25% |

TABLE III. AVERAGE NUMBER OF SUCCESSFUL TASKS IN 30 SECONDS

| Method | Sim | Real |
|---|---|---|
| CSR (2 steps)+DRG (Rotate Z) | **14.4 (30Hz)** | **7.4 (10Hz)** |
| NYU-RL (Rotate Z) | — | 0.4 (30Hz) |
| OpenAI (Rotate XYZ) | 8.6 (24Hz) | 3.6 (12Hz) |

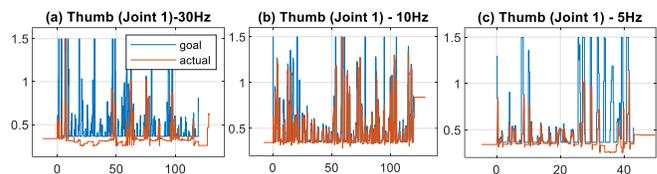

Fig. 5. The following ability of the first joint of the thumb in control frequencies (a) 30Hz, (b) 10Hz, and (c) 5Hz. The robot has the best following ability in 10Hz. In 30Hz, the command is too fast to follow because the speed of the servo motor is limited. In 5Hz, the command is too far from the current position, so the robot cannot reach the target in time.

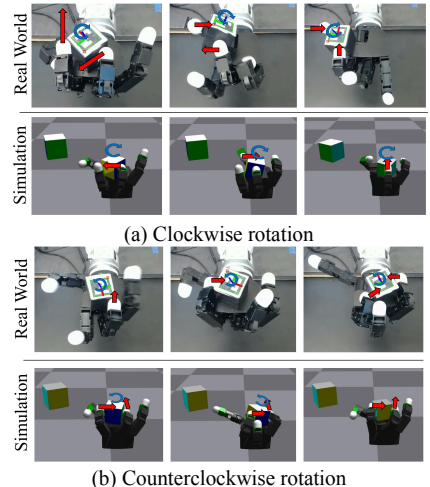

(a) Clockwise rotation

(b) Counterclockwise rotation

Fig. 6. (a) Clockwise rotation (b) Counterclockwise rotation.

Table III compares the CSR (2 steps) + DRG and the performance of similar tasks from the literature OpenAI [6] and NYU-RL [35]. OpenAI used the AAC method with one-step curtail feature reduction from the critic to the actor. NYU-RL learned from human demonstrations with Demonstration Augmented Policy Gradient, where actor and critic have the same feature space. We compare the number of successful cases in 30 seconds. The supplemental video provides more visualized comparisons. It should be noted that OpenAI used images as the input, and the policy was trained with supercomputers with 100 years of experience data. NYU-RL used the same low-dimensional input as ours and was trained on similar computational resources. OpenAI's task is to rotate the object along the XYZ axis and control a Shadow Hand with 24 degrees of freedom, which is more difficult than the Z axes rotation task with the Allegro in our approach and NYU-RL's. The performance of NYU-RL in simulation is not shown because the data is not presented. The control frequencies are not strictly the same due to the different setups, but all reduced control frequencies when deploying to the real world. Due to the modeling discrepancy, the simulation cannot perfectly mimic the physical interaction and dynamics of the object in the real world. Thus, the task performance dropped after the transfer. The results show that CSR (2 steps) + DRG achieved more task success than OpenAI and NYU-RL, showing the promise of our method. OpenAI has a lower number of successes because its task is more difficult. NYU-RL has fewer successes because it learns from human demonstrations, which were intentionally slowed down to improve manipulation stability.

### C. Manipulation Behavior Analysis between Simulation and Real World

This section uses the 10 Hz CSR (2 steps) + DRG as the case study to analyze the manipulation behaviors in simulation and the real world. Fig 6 shows the clockwise and counterclockwise rotations, depending on the shortest distance between the object's current and target positions. The frames are picked to show the typical finger cooperation. In clockwise rotation (Fig. 6a), the robot tends to rotate with the thumb, index, and middle finger or with the index, middle, and ring fingers. In counterclockwise rotation, the index and middle fingers apply the periodic rotation movement, and the thumb and the ring finger prevent the object from falling. Interestingly, we notice that different rotation patterns have different performance reductions when transferred to the real world. Generally, clockwise is more difficult than counterclockwise. The reason is that the Allegro hand in the experiment is a right hand, which is naturally better in the counterclockwise rotation. In the clockwise direction, the rotation relies on the impulse from the edge fingers (thumb or ring, Fig. 6a), making the object freely rotate on the palm, but is unstable compared to the counterclockwise rotation (Fig, 6b), where the index finger and middle finger will adjust the object's position to prepare it for the next impulse. In summary, robot hand structures and different manipulation behaviors will affect the performance when transferring the policy from the simulation to the real world, which is worth further study in the future.

### VI. CONCLUSION

This work incorporated the low-dimensional, rich feature space for policy training in simulation for in-hand manipulation tasks. We proposed the CSR method to generate a curriculum for the actor to discard the undesired features in the real world for effortless policy transfer. We developed the DRG method to remove the existing policy output and input dependencies and avoid creating new ones by generating random signals to replace the reduced feature signals with a deep random network. The simulation and real-world evaluations prove the effectiveness of the proposed method. Our future work will study the potential of the proposed methods in different in-hand manipulation tasks and further refine the methods with automatic curriculum generation.